\documentclass[10pt,leqno]{article}
\usepackage[utf8]{inputenc}
\usepackage[T1]{fontenc}
\usepackage{authblk}            
\usepackage{graphicx,amssymb,amsmath,amsthm,csquotes}
\usepackage{xcolor,paralist,hyperref,titlesec,fancyhdr,etoolbox}
\usepackage[margin=1in]{geometry}
\usepackage{lipsum}
\usepackage{multirow}
\usepackage{makecell}
\usepackage{adjustbox}
\usepackage{booktabs}
\usepackage{tabularx}
\usepackage{array}

\usepackage{todonotes}

\hypersetup{colorlinks=true,linkcolor=black,filecolor=black,urlcolor=black}
\usepackage[backend=biber, style=ieee]{biblatex} 
\titleformat{\section}[display]{\normalfont\huge\bfseries\centering}{\thesection}{10pt}{\Large}
\titlespacing*{\section}{0pt}{0ex}{0ex}

\usepackage{array} 
\newcolumntype{C}[1]{>{\centering\arraybackslash}m{#1}}

\newcommand\blfootnote[1]{%
  \begingroup
  \renewcommand\thefootnote{}\footnote{#1}%
  \addtocounter{footnote}{-1}%
  \endgroup
}

\title{A Multi-Stage Validation Framework for Trustworthy Large-scale Clinical Information Extraction using Large Language Models}

\author[1]{Maria Mahbub\thanks{Corresponding author: \texttt{mahbubm@ornl.gov}}}
\author[2]{Gregory M. Dams}
\author[1]{Josh Arnold}
\author[1]{Caitlin Rizy}
\author[1]{Sudarshan Srinivasan}
\author[2,3]{Elliot M. Fielstein}
\author[4]{Minu A. Aghevli}
\author[5]{Kamonica L. Craig}
\author[2]{Elizabeth M. Oliva}
\author[6,7]{Joseph Erdos}
\author[2]{Jodie Trafton}
\author[1]{Ioana Danciu}

\affil[1]{Oak Ridge National Laboratory, Oak Ridge, TN, USA}
\affil[2]{Program Evaluation and Resource Center, Office of Mental Health and Office of Suicide Prevention, Department of Veterans Affairs, Menlo Park, CA, USA}
\affil[3]{Vanderbilt University Medical Center, Nashville, TN, USA}
\affil[4]{VA Maryland Health Care System, Baltimore, MD, USA}
\affil[5]{VA Desert Pacific Healthcare Network, Long Beach, CA, USA}
\affil[6]{VA Connecticut Health Care System, West Haven, CT, USA}
\affil[7]{Yale School of Medicine, New Haven, CT, USA}

\date{} 

\begin{document}

\maketitle

\begin{abstract}
Large language models (LLMs) show promise for extracting clinically meaningful information from unstructured health records, yet their translation into real-world settings is constrained by the lack of scalable and trustworthy validation approaches.
Conventional evaluation methods rely heavily on annotation-intensive reference standards or incomplete structured data, limiting feasibility at population scale.
We propose a multi-stage validation framework for LLM-based clinical information extraction that enables rigorous assessment under weak supervision.
The framework integrates prompt calibration, rule-based plausibility filtering, semantic grounding assessment, targeted confirmatory evaluation using an independent higher-capacity judge LLM, selective expert review, and external predictive validity analysis to quantify uncertainty and characterize error modes without exhaustive manual annotation.
We applied this framework to extraction of substance use disorder (SUD) diagnoses across 11 substance categories from 919,783 clinical notes.
Rule-based filtering and semantic grounding removed 14.59\% of LLM-positive extractions that were unsupported, irrelevant, or structurally implausible.
For high-uncertainty cases (i.e., LLM-positive but unsupported by structured data), the judge LLM's assessments showed substantial agreement with subject matter expert review (Gwet's AC1=0.80).
Using judge-evaluated outputs as references, the primary LLM achieved an F1 score of 0.80 under relaxed matching criteria.
LLM-extracted SUD diagnoses also predicted subsequent engagement in SUD specialty care more accurately than structured-data baselines (area under the receiver operating characteristic curve=0.80).
These findings demonstrate that scalable, trustworthy deployment of LLM-based clinical information extraction is feasible without annotation-intensive evaluation.
By unifying complementary validation strategies, this work presents a generalizable human-in-the-loop framework that produces scalable, consistent validation of LLM-based clinical information extraction while using targeted expert review to verify automated assessments.



\end{abstract}

\section{Introduction}
\blfootnote{Notice: This manuscript has been authored by UT-Battelle, LLC, under contract DE-AC05-00OR22725 with the US Department of Energy (DOE). The US government retains and the publisher, by accepting the article for publication, acknowledges that the US government retains a nonexclusive, paid-up, irrevocable, worldwide license to publish or reproduce the published form of this manuscript, or allow others to do so, for US government purposes. DOE will provide public access to these results of federally sponsored research in accordance with the DOE Public Access Plan (\url{https://www.energy.gov/doe-public-access-plan}).}

Large language models (LLMs) have demonstrated remarkable capability for extracting clinical information from unstructured electronic health record text \cite{agrawal2022large}.
Extracting substance use disorder (SUD) diagnoses, as defined by the DSM-5 (The Diagnostic and Statistical Manual of Mental Disorders, Fifth Edition; APA, 2013), represents a particularly important application of these capabilities \cite{apa2013diagnostic}.
SUD is a major public health concern, contributing to elevated risks of overdose, hospitalization, and treatment disengagement \cite{abuse2024key,degenhardt2018global}.
While structured administrative data typically record only diagnostic codes, clinical narratives document nuanced clinical assessments of substance use patterns and severity, which are central to risk stratification, treatment planning, and population health surveillance \cite{perlis2012clinical}.
Recent advances in LLMs, such as Llama \cite{dubey2024llama}, GPT-4 \cite{achiam2023gpt}, Claude \cite{anthropic2024claude}, and domain-adapted variants such as Med-PaML \cite{singhal2023large}, have shown near-expert performance for such extraction tasks when evaluated on small, carefully annotated datasets, substantially reducing reliance on brittle rule-based systems and labor-intensive feature engineering \cite{thirunavukarasu2023large}.
Prior work using transformer-based models and generative LLMs has shown promising performance for identifying substance use mentions \cite{wang2015automated}, classifying addiction severity \cite{mahbub2025decoding}, and extracting related social and behavioral context from clinical notes \cite{wang2025extracting}. These studies establish technical feasibility and highlight the potential of LLMs to recover clinically meaningful information that is systematically under-represented in structured data.

Despite these advances, validation becomes the dominant challenge when moving from proof-of-concept to operational deployment. 
Traditional clinical NLP validation assumes the availability of expert-annotated ground truth, against which model outputs are compared using precision, recall, and F1 score.
While this paradigm has enabled rigorous benchmarking in shared tasks and early-stage research, it does not scale to million-note corpora.
Expert annotation is expensive, time-intensive, and constrained by clinician availability, and validation would need to be repeated for each new extraction task, model update, or healthcare system \cite{goel2023llms,wei2018clinical}. 
As a result, many LLM-based clinical NLP systems stall at the pilot stage, despite demonstrating strong performance on small validation sets. 

Validating clinical information extraction systems typically assumes the existence of a reliable ground truth. However, this assumption fails for SUD diagnoses.
Administrative coding systems like ICD-10 \cite{ICD} serve as validation proxies \cite{peng2018coding} but systematically under-represent substance use behavior, SUD diagnoses and clinical complexity due to restrictive coding rules for uncertain or untreated conditions, gaps in clinical SUD expertise, and clinician avoidance of SUD diagnosis due to stigma \cite{atolagbe2021coding}. Discordance between coded diagnoses and narrative documentation is substantial; administrative data frequently miss clinically documented conditions or collapse severity into coarse categories \cite{scherrer2023validating,hurley2025diagnosis}. 
When an LLM extracts evidence of severe or moderate SUD from notes not reflected in ICD-10 encounter coding, traditional validation cannot adjudicate whether the model is incorrect or the structured data is under-coded. This lack of scalable, trustworthy validation acts as a barrier to the deployment of LLM-based SUD extraction.

To address these challenges, we review progress in four pivotal areas: (1) validation approaches for clinical NLP, (2) prompt engineering and calibration, (3) predictive validity assessment, and (4) NLP methods for substance use disorder extraction.

\paragraph{LLM Validation Approaches for Clinical NLP}
The dominant paradigm for evaluating clinical NLP relies on expert-annotated datasets and standard metrics like precision and F1 score \cite{yang2022large,builtjes2025leveraging}. While enabling rigorous benchmarking, this paradigm does not scale to operational corpora.
Emerging approaches use separate LLMs as independent evaluators (LLM-as-a-judge) to assess output quality without exhaustive human annotation. LLM-based judges achieve strong inter-rater reliability with human evaluators for clinical summarization, reducing evaluation time while maintaining concordance \cite{croxford2025automating, he2025llm}. Frameworks for assessing clinical safety and hallucination rates have also established LLM-as-a-judge as a scalable complement to clinician review \cite{asgari2025framework}. These studies confirm the feasibility of model-based evaluation but focus primarily on summarization rather than extraction fidelity.
A related challenge is detecting hallucinations \cite{anh2025survey}. Fabricated diagnoses or findings in clinical applications propagate errors through downstream workflows \cite{pandit2025medhallu,kim2025medical}. The MedHallu benchmark highlights the difficulty of this problem, with state-of-the-art models showing limited performance in detecting subtle medical hallucinations \cite{pandit2025medhallu}. Hybrid frameworks combining information-theoretic classifiers with structured knowledge bases can reduce hallucination rates in clinical question answering \cite{garcia2025trustworthy}. Complementary taxonomies for faithfulness errors and automated assessment methods provide alternatives to manual fact-checking \cite{vishwanath2024faithfulness}.
Most existing approaches target single failure modes and lack evaluation at operational scale. Our framework combines multiple validation signals to balance rigor with efficiency.

\paragraph{Prompt Engineering and Calibration}
LLM performance on clinical tasks is highly sensitive to prompt design. Task-specific prompt tailoring is critical; comprehensive evaluations across tasks like biomedical evidence extraction indicate that chain-of-thought prompting is among the most effective strategies \cite{sivarajkumar2024empirical}. A scoping review of prompt engineering in medicine identified prompt design as the prevalent paradigm, with chain-of-thought reasoning the most frequently adopted technique \cite{zaghir2024prompt}.
Prompt calibration, optimizing content and structure using labeled examples, improves stability. Task-specific frameworks for clinical named entity recognition incorporating annotation guidelines and error analysis instructions through few-shot learning have shown success \cite{zhang2025improving}. Yet, prompt sensitivity remains a challenge. While scaling to larger parameters improves stability, even capable LLMs remain sensitive to prompt formulation, where minor variations in information ordering can significantly degrade clinical performance \cite{hager2024evaluation}.  This underscores the necessity of systematic prompt calibration prior to deployment. Tutorials on clinical prompt engineering emphasize that optimal strategies (zero-shot, few-shot, chain-of-thought) depend on task complexity and computational resources \cite{liu2025prompt}. Carefully designed prompts also improve alignment with evidence-based clinical guidelines \cite{li2025streamlining}.
Iterative prompt refinement combined with structured output templates improves the reliability of extraction from clinical narratives \cite{srivastava2025medpromptextract}. We build on these findings by integrating prompt calibration as a pre-extraction stage to reduce error propagation.

\paragraph{Predictive Validity Assessment}
Predictive validity, assessing whether measurements predict relevant outcomes, has deep roots in clinical psychology and psychometrics. Researchers established construct validity theory, emphasizing that valid measures should predict theoretically related outcomes \cite{cronbach1955construct}.
Across psychiatric conditions, symptom severity measures demonstrate predictive validity for clinically meaningful outcomes. Depression severity scales such as the PHQ-9 and Beck Depression Inventory predict suicide attempts, hospitalization, treatment response, and functional impairment \cite{kroenke2001phq,simon2017between}. Similarly, substance use severity measures. including the Addiction Severity Index, AUDIT, and DSM-5 SUD severity criteria, predict treatment retention, relapse, healthcare utilization, and mortality \cite{rikoon2006predicting,conigrave1995predictive,conigrave1995audit,hasin2013dsm}.
These findings establish the broader principle that valid clinical assessments should predict downstream clinical events in theoretically consistent ways. Accordingly, if LLM-extracted SUD diagnoses reflect true clinical diagnoses, they should predict future SUD-related events.
However, predictive validity has rarely been used as a primary validation strategy for natural language processing systems. Instead, most studies frame outcome prediction as a downstream application rather than as evidence of construct validity. This gap highlights the need for systematic frameworks that leverage predictive validity as a scalable form of external validation, particularly in settings where manual annotation is infeasible.

\paragraph{NLP for Substance Use Disorder Extraction}
Extracting substance use information from clinical text is a longstanding challenge. Early work relied on rule-based and traditional machine learning. NLP systems successfully detected alcohol, drug, and nicotine use attributes from clinical notes \cite{wang2015automated}. Convolutional neural networks applied to electronic health record (EHR) notes achieved strong discriminative performance for substance use screening \cite{afshar2022development}, and real-time deployment of these tools demonstrated operational feasibility in hospital settings \cite{afshar2023deployment}. Classifiers for pediatric populations have also achieved high performance across multiple substance types \cite{ni2021automated}.
Recent approaches leverage zero-shot and few-shot LLM capabilities. Flan-T5 models effectively extract SUD severity specifiers from Veterans Affairs (VA) notes using zero-shot learning, addressing the granularity gap between ICD-10 codes and DSM-5 criteria in unstructured text \cite{mahbub2025decoding}. Features extracted via LLMs improve prediction of treatment retention on medications for opioid use disorder when combined with structured EHR data, demonstrating the value of narrative-derived information \cite{nateghi2025predicting}. More broadly, LLM applications have targeted diverse clinical domains including oncology \cite{chen2025large}, radiology \cite{reichenpfader2024scoping}, and social determinants of health \cite{wang2025extracting}, though most evaluations remain limited to small annotated datasets.

We extend this literature by applying LLM-based extraction to SUD diagnoses at million-note scale and proposing a comprehensive validation framework that addresses trustworthiness challenges when reference standards are unavailable.
While individual validation components have been explored in isolation, their integration into a unified, outcome-anchored evaluation pipeline at million-note scale represents a novel contribution.
We organize validation of LLM-based clinical extraction into two complementary stages: pre-extraction reliability and post-extraction validation. Pre-extraction reliability focuses on ensuring consistent and interpretable model behavior through prompt calibration and controlled inference settings. Post-extraction validation addresses the more challenging problem of determining whether extracted information is correct and clinically meaningful in the absence of definitive ground truth.
Together, these stages provide layered evidence of extraction correctness, reducing reliance on any single validation signal.
The rest of this paper is organized as follows: Section \ref{sec:method} describes the methods; Section \ref{sec:result} presents results for each validation stage and discusses implications and limitations with possible future directions; and Section \ref{sec:conclusion} concludes.

\section{Methods}
\label{sec:method}
\begin{figure}[!htbp]
    \centering
    \includegraphics[width=\linewidth]{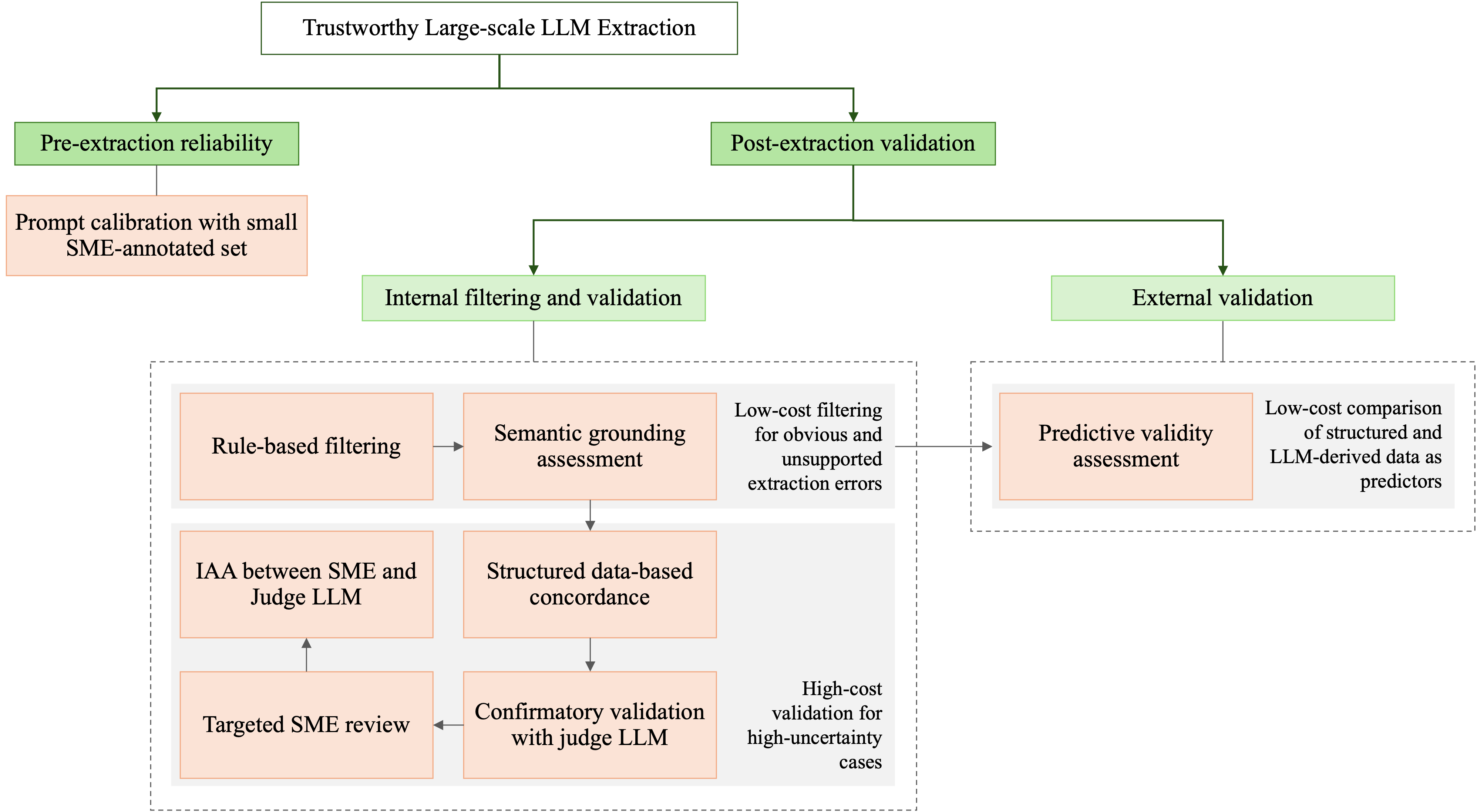}
    \caption{A multi-stage framework for trustworthy large-scale LLM extraction of clinical information}
    \label{fig:framework}
\end{figure}

The proposed multi-stage framework is designed to assess and improve the trustworthiness of LLM-based information extraction through modular pre-extraction and post-extraction components.
As shown in Figure \ref{fig:framework}, pre-extraction reliability was addressed through prompt calibration using a small set of annotated notes, with the goal of promoting more consistent extraction behavior on new data.
Post-extraction validation consisted of internal (extraction-level) and external (outcome-level) components that can be applied independently or in combination depending on the information extraction task, data availability, and computational feasibility.
Internal filtering removed clearly unsupported extractions using rule-based criteria and semantic grounding assessment.
Internal validation then evaluated retained extractions using complementary signals, including concordance with structured data, limited confirmatory assessment using a judge LLM applied to a small subset of high-uncertainty cases, and targeted review by a clinical psychologist subject matter expert (SME) to assess the reliability of the judge LLM.
External validation assessed predictive validity by testing whether LLM-extracted data exhibited outcome-relevant behavior comparable to or exceeding that of related structured data.
By integrating textual support, structured data alignment, expert judgment, and downstream clinical relevance, this layered yet flexible framework enables systematic assessment of extraction trustworthiness in large-scale settings.

The remainder of this section describes the implementation of each framework component, the datasets used in the study, the experimental setup, and the evaluation metrics used to assess effectiveness of each component.

\subsection{Pre-extraction Prompt Calibration}
\label{sec:prompt_cal}
To improve extraction quality, we performed systematic prompt optimization using a small, SME-annotated dataset of clinical notes, extended from our prior work \cite{mahbub2025decoding}. By evaluating multiple prompting strategies before deployment, we aimed to reduce error propagation through the extraction pipeline.

\begin{figure}[!htbp]
    \centering
    \includegraphics[width=\linewidth]{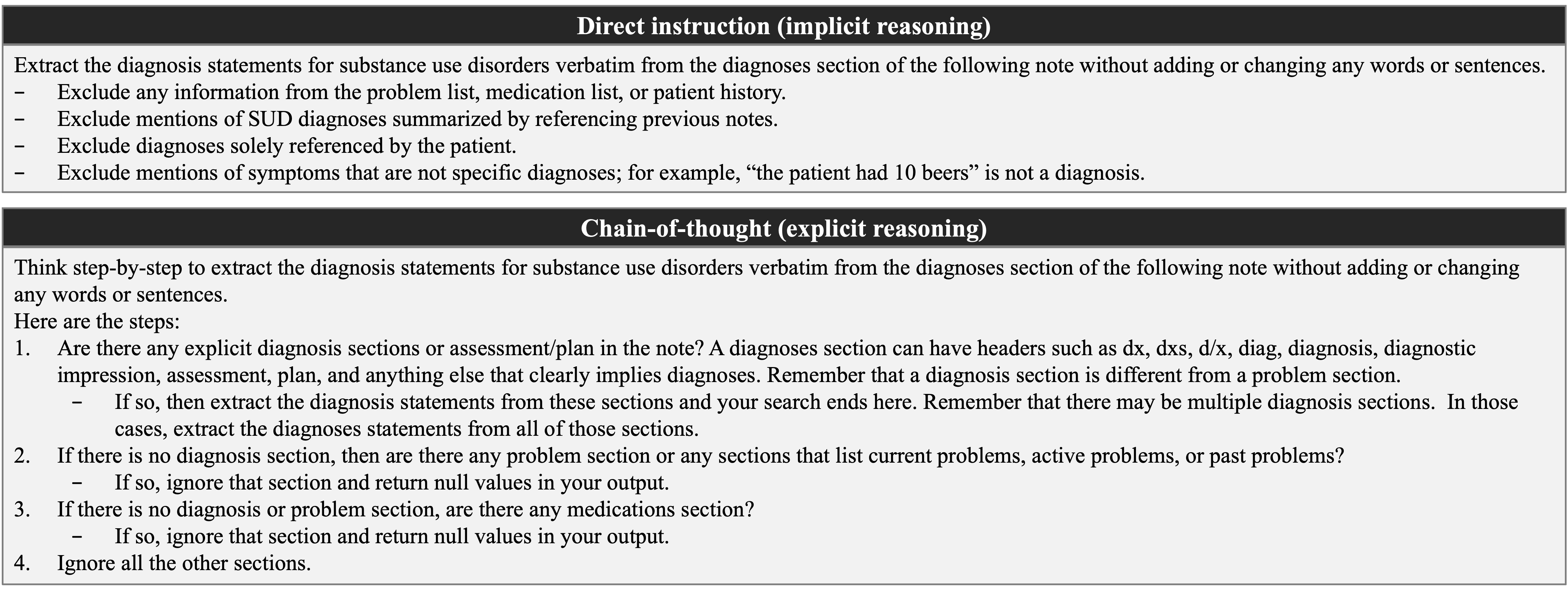}
    \caption{Direct vs chain-of-thought prompting to extract SUD diagnoses information from clinical notes.}
    \label{fig:prompt}
\end{figure}

We evaluated zero-shot, one-shot, and two-shot prompting strategies by crafting prompts that optionally accepted note–annotation pairs as in-context examples for the LLM. Note–annotation pairs were selected from the annotated dataset using a fixed-mode strategy, which ranked notes by the total number of SUDs labeled per note and selected the top-k most informative examples from the full dataset. For the best-performing n-shot configuration, we additionally compared direct prompting with chain-of-thought prompting.
Direct prompts specify the desired output without explicit reasoning, whereas chain-of-thought prompts explicitly guide the model through intermediate reasoning steps before producing an answer.
The chain-of-thought prompt was structured to address recurring errors observed during iterative refinement of the direct prompt.
The strategy that achieved the strongest performance on the annotated data was subsequently deployed at scale.
The differences between direct and chain-of-thought prompts are provided in Figure \ref{fig:prompt}.
All prompting strategies followed strict instruction-constrained, schema-guided, and source-grounded prompting, where the LLM was explicitly restricted to extracting only information stated in the clinical text (no inference or normalization), outputs were forced into a predefined structured json schema, and every extracted information was directly linked to its originating clinical note for traceability.

\paragraph{Evaluation Metrics:}
LLM extraction performance on the annotated set was evaluated using two string-matching criteria, strict and relaxed, across three metrics: precision, recall, and F1 score \cite{manning2008introduction}. Strict matching required exact character-level agreement between generated text and ground truth and was scored binarily (F1 = 1 for exact matches; 0 otherwise). Relaxed matching measured token-level overlap and computed precision, recall, and F1 scores on a continuous scale from 0 to 1. Macro-averaged metrics were reported across the set, with true positives defined as overlapping tokens, false negatives as ground-truth-only tokens, and false positives as generated-only tokens.


\subsection{Post-extraction Rule-based Filtering}
As the first step of post-extraction validation, we applied rule-based filtering to efficiently identify and remove outputs irrelevant to the model query or obvious extraction errors, prior to more computationally intensive validation stages.
Rule-based extraction from raw clinical notes is challenging because clinically relevant concepts may appear in multiple contexts that do not carry equivalent meaning for information extraction \cite{savova2010mayo}. For example, terms such as ``alcohol use disorder'' may appear in past problem lists, past medical history, templated screening sections, or copied-forward text, even when the current note does not document provider-assessed SUD diagnosis or severity (i.e., SUD diagnosis specifiers). As a result, simple pattern matching cannot reliably distinguish between these contexts, leading to high false-positive rates when applied directly to unstructured notes.

In contrast, applying rule-based checks to LLM-extracted outputs provides an effective validation layer, as the rules operate on concise text spans generated by the LLM rather than full clinical documents. In this setting, computational rules need only verify the internal consistency and plausibility of a small, focused text string, for example, detecting negation or irrelevant phrases, rather than resolving complex document structure. This enables efficient identification of obvious extraction errors at scale.
We have also considered using rule-based filtering before extraction, which could potentially reduce scale and computational burden; however, this would shift complexity pre-extraction and compromise the framework's goal of preserving recall while deferring precision control to post-extraction validation.

For SUD diagnoses, and with input from an SME, we curated two sets of domain-specific phrases: inclusion phrases, of which at least one must be present for an extraction to be considered valid, and exclusion phrases, none of which may be present. Extractions lacking all inclusion phrases or containing any exclusion phrase were flagged as LLM extraction errors. The full phrase lists are provided in Table~\ref{tab:rule}.
These phrase-based rules were intentionally designed to be conservative, prioritizing high precision over recall due to the asymmetric consequences of model errors in a clinical context. When the model fails to identify a diagnosis, existing structured data (e.g., ICD-10 codes) remains unchanged and continues to serve as the operational ground truth for downstream use. In contrast, when the model proposes a new diagnosis or modification that is not explicitly present in the structured fields, the output is necessarily inferential and may introduce clinically meaningful downstream effects. In this setting, false-positive inferences are more harmful than false negatives, as they risk introducing unsupported or erroneous diagnostic information. Accordingly, we require that any proposed addition or update to structured diagnoses be supported by clear, explicit, and unambiguous evidence in the source text. This conservative design choice minimizes the risk of erroneous alterations to structured clinical data while preserving safety in deployment.

\begin{table}[!htbp]
\caption{Inclusion and exclusion phrases for post-extraction rule-based filtering.}
\label{tab:rule}
\centering

\begin{adjustbox}{max width=\linewidth}
\begin{tabularx}{\linewidth}{XX}
\toprule
Inclusion phrases & Exclusion phrases \\
\midrule
`dep\textbackslash s*', `d/o', `disorder', `abuse', `ud\textbackslash s*', `use\textbackslash s*do\textbackslash s*', `do\textbackslash s*', `sev', `mild', `moderate', `unspec', `user', `use ', `use dis', `diag', `dx', `dgx', `detox', `intox', `withdraw', `remiss', `addict', `w/d', and substance list from \cite{mahbub2025decoding}
&
`history', `hx', `h/o', `ho\textbackslash s+',  `uses', `has been', `audit-c', `ciwa', `drink', `pack', `year old', `y/o', `admission', `years', `ago', `blood', `lozenge', `patch', `gum', `mg\textbackslash s+', `encourage', `offer', `avoid', `treat', `stop', `recommend', `discuss', `advise', `refuse', `not interested', `address', `should', `r/o', `ro ', `rule out', `deny', `denies', `denied', `quit' \\
\bottomrule
\end{tabularx}
\end{adjustbox}
\end{table}

\subsection{Post-extraction Semantic Grounding Assessment}
Following rule-based filtering, we applied semantic grounding assessment to the remaining LLM-extracted outputs to evaluate whether extracted SUD severity information was supported by evidence in the source clinical note (i.e., to evaluate the risk of hallucinated vs paraphrased outputs). This stage targets a complementary failure mode, plausible but unsupported generation, by identifying extractions whose content cannot be semantically traced to the original documentation, without requiring manual review.

For each LLM-generated extraction, semantic alignment with the corresponding clinical note was assessed using embedding-based cosine similarity.
Text embeddings were computed using a lightweight SentenceTransformer model \cite{reimers2019sentence} for both the extracted output and overlapping sliding windows applied across the full clinical note. Sliding windows were used because the evidence supporting SUD diagnoses is often localized to short spans of text (e.g., assessment or diagnosis sections) rather than distributed uniformly across long notes.
The window length was set to match the length of the extracted text and advanced across the note with a stride of one character to enable fine-grained localization of supporting evidence.

Cosine similarity was computed between the extraction embedding and each window embedding, and the maximum similarity across all windows was retained as the grounding score. This score represents the strongest semantic correspondence between the generated content and any portion of the note, ensuring that each extraction is evaluated against its most relevant textual context.
Cosine similarity was selected because it captures semantic relatedness beyond exact lexical overlap and is robust to paraphrasing, allowing grounded extractions to be identified even when the LLM summarizes or reformulates clinical language rather than reproducing it verbatim.
Extractions with grounding scores below a defined threshold were flagged as poorly grounded, indicating insufficient semantic support in the source text and elevated risk of hallucination. The similarity threshold was selected empirically based on inspection of the grounding score distribution.


\paragraph{Evaluation Metrics:}
As an evaluation metric for both rule-based filtering and semantic grounding assessment, we report the count of flagged extractions, defined as the total number of LLM+ outputs rejected by these validation stages. An LLM+ output is defined as any instance in which the LLM returns a non-empty string extracted from the source note. Because these checks prioritize precision, the flagged extractions serve as an estimate of the proportion of LLM+ outputs that are likely spurious. Accordingly, a high count of flagged extractions would indicate that a substantial portion of raw LLM extractions would constitute false positives in the absence of validation, underscoring the importance of these early-stage checks.

\subsection{Post-extraction Confirmatory Validation with Judge LLM}
Following rule-based filtering and semantic grounding assessment, we applied confirmatory validation using a larger judge LLM. Because larger models require substantially greater computational resources and are often infeasible for routine clinical deployment, this stage was applied only to a targeted subset of extractions in which disagreement with relevant structured data was most clinically consequential. In our framework, judge model confirmation functions as a targeted deep-validation stage, invoked only when lower-cost automated checks are insufficient to resolve clinically meaningful disagreement. By restricting this step to a small, high-uncertainty subset of cases, we obtained stronger validation signals while maintaining computational feasibility at scale.
For triggered cases, the judge LLM independently evaluated each extraction in the context of the source clinical note. The model was prompted to assess whether the extracted content is supported by documented clinical assessments and relevant contextual indicators. Agreement between the primary extraction model and the judge LLM serves as an additional validation signal, enabling independent adjudication of challenging cases without requiring large-scale manual annotation.

\begin{figure}[!htbp]
    \centering
    \includegraphics[width=\linewidth]{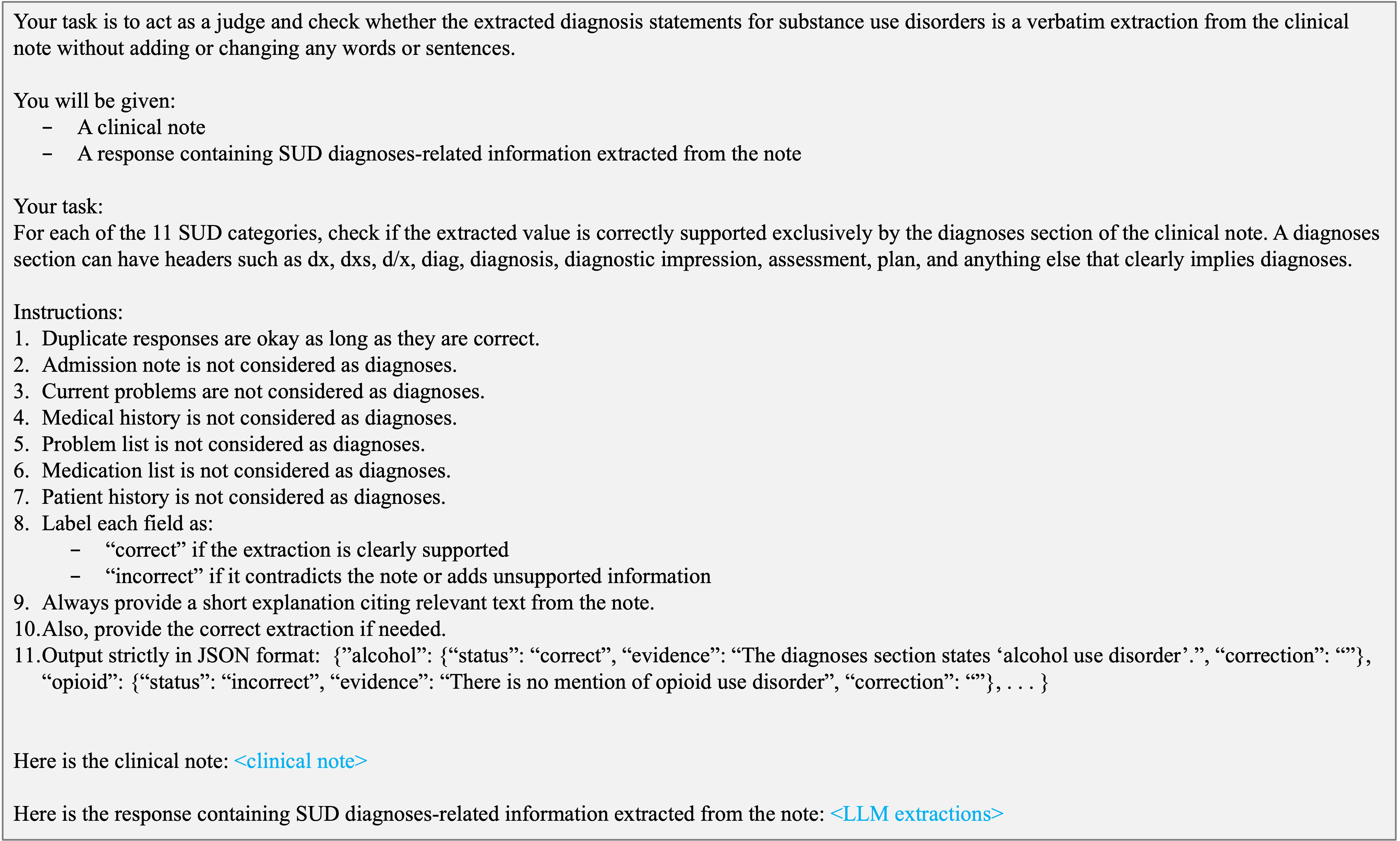}
    \caption{Prompt for judge LLM}
    \label{fig:judge_prompt}
\end{figure}

For SUD diagnoses, confirmatory validation was triggered only when the LLM extracted evidence of a diagnosis in the absence of a corresponding ICD-10 code. While semantic grounding assessment ensures that extractions are attributable to the source note, it cannot determine whether a mention represents a current provider assigned diagnosis as opposed to historical or templated content. In clinical settings, false-positive SUD diagnoses are generally considered more harmful than false negatives, as unsupported diagnoses may propagate through downstream clinical workflows, introduce inappropriate interventions, and carry stigmatizing or regulatory consequences. In contrast, false negatives typically represent omissions that can be addressed through subsequent clinical evaluation. The judge prompt is provided in Figure \ref{fig:judge_prompt}.

\paragraph{Evaluation Metrics:}
The judge LLM labeled each extraction as correct or incorrect and, when incorrect, provided a corrected extraction along with supporting evidence. To account for partial correctness, as discussed in Section~\ref{sec:prompt_cal}, we evaluated performance using standard information extraction metrics, including strict and relaxed F1, relaxed precision, and relaxed recall (Section~\ref{sec:prompt_cal}). Reference extractions were defined as follows: for outputs labeled correct by the judge LLM, the original LLM extraction was treated as the reference; for outputs labeled incorrect, the judge-provided correction was used as the reference.

\subsection{Post-extraction SME Review}
The final internal validation stage consists of targeted review by an SME to assess agreement between expert clinical judgment and automated validation signals. Rather than establishing a definitive ground truth for all extractions, this step is intended to evaluate whether the judge LLM's assessments align with human expert interpretation when applied to a subset of high-uncertainty cases.
A random sample of notes from the confirmatory validation subset was reviewed by the SME. For each case, the SME was provided with the source clinical note and the corresponding LLM extraction and asked to determine whether the extraction was supported by the documentation. Independently, the same cases were evaluated by the judge LLM using identical source material.
The SME additionally assessed the quality and appropriateness of the judge LLM's reasoning and decision.

\paragraph{Evaluation Metrics}
Agreement between the SME and the judge LLM was quantified using inter-annotator agreement (IAA) metrics. Specifically, we used Gwet's AC1 \cite{gwet2001handbook}, a chance-corrected coefficient for binary correct/incorrect judgments that is robust to data imbalance.
We selected AC1 over Cohen's Kappa due to the substantial label skew in our dataset, i.e., one outcome (``correct'') occurs far more frequently than the other (``incorrect''), as Kappa is known to underestimate agreement in highly imbalanced settings (the Kappa Paradox) \cite{derksen2024kappa}, potentially yielding low scores despite high observed agreement.
AC1 values range from -1 to 1, where higher values indicate stronger agreement beyond chance, with commonly used qualitative interpretations of poor (<0.40), moderate (0.40-0.60), substantial (0.60-0.80), and near-perfect (>0.80) agreement \cite{landis1977measurement}.
This analysis evaluates the extent to which the automated judge model's assessments align with expert clinical judgment, providing evidence for whether the judge model can serve as a reliable surrogate reviewer at scale.
Gwet's AC1 is preferred in this setting because the evaluation task is expected to be dominated by correct judgments, a condition under which Cohen's kappa is known to underestimate agreement.
High agreement supports the use of the judge LLM as an efficient confirmatory validation mechanism, whereas disagreement highlights residual ambiguity in the extracted information or limitations of automated assessment.
Within our framework, targeted human review therefore functions primarily as a calibration and trust-establishing step for the judge model, rather than as a comprehensive manual annotation effort.

\subsection{Predictive Validity Assessment}
While internal (extraction-level) validation assesses fidelity with respect to source documentation, it does not establish whether extracted signals generalize beyond the source note or correspond to clinically meaningful outcomes. We therefore conducted a predictive validity assessment to evaluate whether LLM-extracted information is associated with downstream clinical events in a manner consistent with established clinical knowledge. Demonstrating such alignment provides an external, outcome-oriented validation of LLM extractions and strengthens confidence that extracted signals reflect clinically meaningful information rather than spurious or stylistic artifacts of documentation.
As a reference, we used the most closely related structured data, previously leveraged to identify high-uncertainty cases, as a baseline for predictive performance. Predictive performance comparable to or exceeding that of the structured-data baseline would indicate that LLM extractions capture outcome-relevant clinical signal and do not introduce unsupported risk indicators.

We employed a simple logistic regression model to predict outcomes derived from structured clinical data. Analyses were conducted across three cohorts: (1) the full study population, to assess overall predictive performance; (2) cases with concordant LLM response and structured data, which served as a reference comparison group; and (3) cases with LLM extractions in the absence of corresponding structured data, to evaluate the clinical relevance of narrative-only information. This stratification enables assessment of whether LLM extractions behave like established clinical signals both overall and in settings where structured data may under-represent documented information. If LLM extractions reflect clinically meaningful signal rather than spurious artifacts, predictive performance in the narrative-only cohort would be expected to approximate that of the concordant cohort.

For SUD diagnoses, predictive validity was assessed by evaluating whether LLM-extracted diagnoses predict subsequent engagement with SUD specialty care, compared with ICD-10–based diagnoses. This outcome reflects downstream clinical recognition, referral, and treatment engagement, providing an external validity check on whether SUD diagnoses extracted from clinical notes in the predictor window are associated with near-term SUD specialty care utilization in the outcome window. For each cohort, three predictor sets were evaluated: (i) SUD diagnoses derived from ICD-10 codes; (ii) SUD diagnoses extracted by the LLM from clinical narratives; and (iii) a combined feature set incorporating both sources.

In the full cohort, predictive models incorporated all SUD categories jointly to assess overall performance across predictor sets. In the narrative-only and concordant cohorts, analyses were restricted to the most prevalent SUD category, as determined by ICD-10 code frequency in the full cohort. This restriction avoided ambiguity arising from inconsistent substance category-specific documentation within individual records and prevented sparsely populated strata that would yield unstable estimates. Focusing on a single, prevalent category ensured consistent cohort definition and robust estimation of predictive performance.

ICD-10–based diagnoses were encoded as binary indicators across SUD categories corresponding to ICD-10 codes F10–F19. LLM-extracted diagnostic information was concatenated across all notes per patient and encoded using term frequency–inverse document frequency (TF–IDF) representations of the extracted text spans.

\paragraph{Evaluation Metrics:}
Predictive performance was evaluated using the area under the receiver operating characteristic curve (AUC), a threshold-independent measure of discrimination that is robust to class imbalance. Comparative performance across predictor sets was used to assess whether LLM-extracted information captured outcome-relevant clinical signal.

\subsection{Datasets}
We used curated clinical data from Veterans Affairs Corporate Data Warehouse (VA CDW). Clinical notes were selected from patient encounters associated with substance use disorder (SUD)–related visits, identified using ICD-10 diagnosis codes beginning with F10, F11, F12, F13, F14, F15, F16, F17, F18, and F19 recorded in January 2023. The resulting dataset comprised 178,467 unique patients with a total of 919,783 clinical notes.

LLM-based extraction targeted 11 SUD categories, consistent with prior work \cite{mahbub2025decoding}. For prompt calibration, we used an extended expert-annotated dataset derived from our prior work \cite{mahbub2025decoding}. This calibration dataset consisted of a random sample of 923 fully identified clinical notes from 685 unique patients, curated from the VA CDW. These notes were used exclusively for prompt development and were not included in post-extraction evaluation analyses.

For the predictive validity assessment, outcomes were defined using structured data indicating whether a patient engaged in SUD specialty care during the observation window spanning February 2023 to April 2023. Engagement in SUD specialty outpatient care was identified using predefined clinic codes, with 51552 (28.89\%) patients meeting the outcome criterion during the observation window.

Based on ICD-10 diagnosis codes in the full cohort, alcohol and nicotine were the most prevalent SUD categories, affecting 55.40\% and 30.93\% of patients, respectively, whereas hallucinogen and inhalant were the least prevalent, each affecting 0.22\% and 0.07\% of patients.
Given this distribution, subsequent analyses focused on alcohol use disorder (AUD), the most prevalent SUD category, to ensure stable cohort definition and reliable estimation of predictive performance.

Patients were stratified into mutually exclusive groups based on concordance between structured and narrative indicators of AUD. This resulted in 144867 patients with both ICD-10–based AUD diagnoses and corresponding LLM-extracted evidence (concordant cohort), and 7014 patients with LLM-extracted AUD evidence in the absence of ICD-10 diagnoses (narrative-only cohort).
Outcome prevalence differed across these cohorts, with 32.19\% of patients in the narrative-only cohort and 30.83\% in the concordant cohort engaging in SUD specialty care during the outcome window.

For model development and testing, data for each cohort were partitioned into training (70\%), validation (10\%), and test (20\%) sets at the patient level to prevent information leakage across splits. All feature engineering and model selection were performed using the training and validation sets, with final performance evaluation conducted exclusively on the held-out test set. This split strategy was applied consistently across models using ICD-10 features, LLM-extracted features, and combined feature sets to ensure fair comparison.

\paragraph{Ethics:} This project was conducted as a national quality improvement effort to improve care for veterans with substance use being treated in the VA. Models were designed to be implemented into VA decision support systems, and are not expected to be generalizable or valid for application outside of notes from the VA Computerized Patient Record System. As such, this work is considered non-research by VA (as per ProgramGuide-1200-21-VHA-Operations-Activities.pdf (va.gov)). However, Oak Ridge National Laboratory (ORNL) required additional oversight of this VA clinical quality improvement project as local standard practice for all uses of patient medical record data within their institution, with approval of the project by the ORNL IRB.
Data access and analysis were performed within a secure clinical data environment, and no protected health information was exported outside the approved research setting.

\subsection{Experimental Setup}
The LLMs used in this study are selected to illustrate the proposed validation framework; the framework itself is model-agnostic and can be applied to other instruction-tuned LLMs with comparable capabilities.

The information extraction was performed using Llama-3.1-8B-Instruct \cite{dubey2024llama} as the primary inference model. This model was selected to balance extraction quality, inference efficiency, and deployment feasibility at million-note scale.
Because the primary contribution of this work is the proposed validation framework rather than model benchmarking, the selected model serves as a representative open-source instruction-following model and can be readily replaced with newer alternatives.
Compared with larger proprietary models, Llama-3.1-8B-Instruct provides strong instruction-following performance while remaining computationally tractable. Its open-weight availability also enables reproducibility and controlled deployment within secure clinical computing environments, which is essential for processing protected health information (PHI).

Confirmatory validation was performed using gpt-oss-20b \cite{agarwal2025gpt}, a higher-capacity open-weight language model deployed as an independent judge. This model was selected to provide a stronger reasoning signal than the primary extraction model while maintaining architectural independence, thereby reducing the risk of shared failure modes and self-consistency bias. 

All LLM inference and validation stages were conducted on a high-performance computing environment equipped with eight NVIDIA A100 GPUs (81.92 GB memory each).

\section{Results and Discussion}
\label{sec:result}
This section reports the results of experiments conducted to evaluate the proposed framework for enhancing the trustworthiness of LLM-extracted information from clinical notes.

\subsection{Pre-extraction Reliability}
Table~\ref{tab:prompt_cal} summarizes extraction performance on the annotated dataset across zero-shot, one-shot, and two-shot prompting strategies, evaluated under both strict and relaxed matching criteria for direct prompting. Zero-shot prompting achieved the highest extraction performance under both strict (F1 = 0.8203) and relaxed (F1 = 0.8647) criteria, outperforming one-shot and two-shot strategies.
One plausible explanation is that in-context examples drawn from a relatively small, SUD-dense subset of notes introduced distributional bias, which may have limited generalization to the broader evaluation set. In contrast, zero-shot prompting avoided this bias while preserving instruction-following performance.
Incorporating chain-of-thought prompting further improved performance in the zero-shot setting, yielding a strict F1 score of 0.8582 and a relaxed F1 score of 0.9004. This improvement likely reflects the model's ability to explicitly reason over multiple pieces of clinical evidence within a note, which is particularly beneficial for capturing nuanced and context-dependent SUD information.
Based on these results, zero-shot chain-of-thought prompting was selected for large-scale deployment.

\begin{table}[!htbp]
\centering
\caption{Prompt calibration results on the annotated dataset comparing zero-, one-, and two-shot prompting under direct and chain-of-thought strategies.}
\begin{tabular}{cccccc}
\toprule
\makecell{Reasoning type} & \makecell{Number of \\shots} & \makecell{Precision \\(relaxed)} & \makecell{Recall \\(relaxed)} & \makecell{F1 \\(relaxed)} & \makecell{F1 \\(strict)} \\ \hline
\multirow{3}{*}{Direct} & 0              & 0.8782                       & 0.8691                    & 0.8647                & 0.8203              \\
                        & 1              & 0.8772                       & 0.8564                    & 0.8584                & 0.8113              \\
                        & 2              & 0.8648                       & 0.8517                    & 0.8514                & 0.8031              \\ \hline
Chain-of-thought        & 0              & 0.9189                       & 0.8973                    & 0.9004                & 0.8582   \\ \bottomrule          
\end{tabular}
\label{tab:prompt_cal}
\end{table}

\subsection{Post-extraction Internal Filtering}
Following LLM extraction, we evaluated the effectiveness of the initial automated filtering stages -- rule-based filtering and semantic grounding assessment -- in identifying low-confidence outputs at scale in the absence of annotations.
LLM extraction identified any SUD diagnoses-related information in 42.41\% of notes, corresponding to 390081 notes in total.
Among 10,117,613 total possible extraction attempts, defined as the product of the total number of clinical notes and the number of SUD categories, 622,551 resulted in non-empty LLM+ extractions.
This highlights the sparsity of clinically relevant SUD mentions across notes and categories presumably due to these patients receiving multiple encounters outside of SUD specialty services.
We also report the count of flagged extractions for each substance category across the dataset. In Table~\ref{tab:internal_filter}, we present LLM+ extraction counts before and after internal filtering to quantify their impact.
As shown in Table~\ref{tab:internal_filter}, pre-filter LLM+ extraction counts varied by substance category, with the highest count of 261037 for alcohol use disorder, as expected.
Elevated count of flagged extractions observed during rule-based filtering indicate that the LLM is extracting content not pertinent to positive SUD diagnoses. Variation in flagged extraction counts across substance categories likely reflects heterogeneity in clinical documentation practices and variability in diagnostic specificity among SUDs.


\begin{table}[!htbp]
\centering
\caption{LLM+ extraction counts and flagged extraction counts from rule-based filtering and semantic grounding assessment by substance category.}
\begin{tabular}{lcccccccc}
\toprule
\multirow{3}{*}{\makecell{Substance category}} & \multirow{3}{*}{\makecell{Pre-filter \\LLM+}} & \multicolumn{6}{c}{\makecell{Count of flagged LLM+ extractions}}                                                                                                                                      & \multicolumn{1}{c}{\multirow{3}{*}{\makecell{Post-filter \\LLM+}}} \\ \cline{3-8}
                                             &                                                               & \multirow{2}{*}{\makecell{Rule-based}} & \multicolumn{5}{c}{\makecell{Semantic grounding}}                                                                                & \multicolumn{1}{c}{}                                                \\ \cline{4-8}
                                             &                                                               &                                                & \makecell{0.50} & \makecell{0.55} & \makecell{0.60} & \makecell{0.65} & \makecell{0.70} & \multicolumn{1}{c}{}                                                \\ \hline
Alcohol                                      & 261037                                                       & 29692                                          & 972                     & 1064                    & 1260                    & 1708                    & 2270                    & 229637                                                             \\
Opioid                                       & 63123                                                        & 14123                                          & 824                     & 855                     & 947                     & 1057                    & 1326                    & 47943                                                              \\
Cannabis                                     & 71777                                                        & 5917                                           & 1065                    & 1093                    & 1150                    & 1315                    & 1579                    & 64545                                                              \\
Cocaine                                      & 65796                                                        & 4066                                           & 873                     & 888                     & 920                     & 1025                    & 1317                    & 60705                                                              \\
Amphetamine                                  & 40992                                                        & 2455                                           & 607                     & 625                     & 640                     & 708                     & 848                     & 37829                                                              \\
Caffeine                                     & 1567                                                         & 480                                            & 400                     & 400                     & 403                     & 406                     & 414                     & 681                                                                \\
Hallucinogen                                 & 2875                                                         & 682                                            & 661                     & 663                     & 671                     & 682                     & 686                     & 1511                                                               \\
Nicotine                                     & 87046                                                        & 16446                                          & 731                     & 764                     & 826                     & 943                     & 1088                    & 69657                                                              \\
Inhalant                                     & 1332                                                         & 334                                            & 400                     & 404                     & 416                     & 425                     & 451                     & 573                                                                \\
Other or unknown                             & 16335                                                        & 4087                                           & 365                     & 374                     & 390                     & 406                     & 442                     & 11842                                                              \\
\makecell{Sedative, hypnotic, \\ or anxiolytic}            & 10671                                                        & 3086                                           & 730                     & 743                     & 753                     & 782                     & 830                     & 6803             \\                                                 
\bottomrule
\end{tabular}
\label{tab:internal_filter}
\end{table}

The flagged extractions were subsequently filtered and the remaining ones were passed onto the next filtering stage, semantic grounding assessment, which further identified extractions that are poorly supported by the clinical notes.
We evaluated across a range of similarity thresholds -- 0.50, 0.55, 0.60, 0.65, 0.70 -- and selected 0.65 as a default operating point for subsequent LLM validation, as it provided conservative filtering while preserving the majority of extractions for subsequent review.
Semantic grounding flagged a disproportionately high fraction of LLM+ extractions for caffeine, hallucinogen, and inhalant use disorders.
Manual inspection of a subset of flagged cases indicated that these elevated counts were primarily driven by very short or non-informative clinical notes, e.g., `consult completed' or `see admission note', in which the LLM generated unsupported extractions across all substance categories in the absence of substantive clinical context. Because these SUDs were rare in the extracted corpus, a small number of notes with such unsupported generations resulted in inflated flag proportions.
The count of flagged extractions within each substance category was stable across the similarity thresholds, indicating robustness to threshold selection.
Collectively, rule-based filtering and semantic grounding removed 14.59\% of total LLM+ extractions across 11 SUD categories.
Table~\ref{tab:internal_filter} summarized the distribution of LLM+ extractions by SUD categories after internal filtering.

\subsection{Post-extraction Internal Validation}
Following rule-based filtering and semantic grounding assessment, LLM-derived SUD diagnoses were evaluated by comparison with ICD-10 diagnostic codes assigned at the encounter level within each substance category. Because ICD-10 codes are assigned per encounter and each encounter may contain multiple clinical notes, LLM-derived diagnoses were aggregated across all notes within an encounter prior to comparison.
Both cases in which the LLM identified a diagnosis (LLM+) and those in which it did not (LLM-) were included, allowing agreement to be assessed across all encounters. For each substance category, agreement rate was quantified as the proportion of encounters in which the LLM output aligned with the corresponding ICD-10 diagnosis (i.e., concordant presence or absence within the category).

Agreement rates were 75.28\% for alcohol, 91.93\% for opioid, 91.77\% for cannabis, 91.24\% for cocaine, 93.58\% for amphetamine, 91.15\% for caffeine, 99.69\% for hallucinogen, 86.26\% for nicotine, 99.91\% for inhalant, 95.46\% for other or unknown, and 98.96\% for sedative, hypnotic, or anxiolytic.
This level of correspondence provides evidence of strong concurrent validity, a standard psychometric indicator describing the extent to which independent measures capture the same underlying construct. The consistently high alignment between LLM outputs and ICD-10 diagnoses across substance categories indicates that the LLM reliably identifies SUD diagnoses, supporting its validity as a diagnostic extraction tool.

For instances with LLM+ extractions but no corresponding ICD-10 diagnoses, a secondary confirmatory review using a judge LLM was applied. Specifically, LLM+ extractions from 80006 clinical notes across 76549 encounters lacked a corresponding ICD-10 diagnosis and met the trigger criteria for judge-based review.
Within this subset, the judge LLM confirmed the majority of primary LLM extractions.
Table~\ref{tab:judge} summarizes the judge LLM's evaluation of extraction validity for all triggered cases. For each substance category, we consider only instances where the primary LLM produced an extraction, excluding cases with no extractions, as our analysis focuses on evaluating the quality of extracted results.
Overall, extracted SUD diagnoses achieved 0.7735 strict F1, 0.7962 relaxed F1, 0.7963 relaxed precision, and 0.8011 relaxed recall scores when evaluated against judge-provided feedback.
Performance varied across substance categories, with strict F1 scores ranging from 0.5761 to 0.8446.
These results indicate strong alignment between the primary LLM and the judge LLM in identifying evidence-supported SUD diagnoses for most substance categories.
The performance scores suggest that a substantial proportion of LLM+, ICD-negative extractions reflect clinically meaningful information present in free-text documentation but not captured by structured diagnostic codes. Broadly this may represent that the provider believes the diagnosis is present but that particular condition was not being actively treated during the encounter or it may represent providers who are making provisional diagnoses based on chart review.
Variation in performance across substances likely reflects differences in documentation practices and the degree to which SUD diagnoses is explicitly recorded in clinical notes.

Disagreements between the judge LLM and the primary LLM largely stemmed from challenges in distinguishing documented SUD diagnoses from other diagnosis-like statements in free-text clinical notes. Manual review showed that most errors resulted from misattributing non-diagnostic mentions, such as templated sections or patient-reported history, as formal diagnoses, or from inferring diagnoses based on use-pattern descriptions (e.g., ``heavy drinker'') without explicit diagnostic documentation. Additional errors primarily involved cases in which the primary LLM failed to identify explicit diagnostic statements. Collectively, these findings highlight the potential of the judge LLM as a scalable confirmatory validation component within the framework, particularly for quantifying agreement and systematically characterizing error patterns.



\begin{table}[!htbp]
\centering
\caption{Information extraction performance for SUD diagnoses as assessed by the judge LLM, considering only instances where the primary LLM produced an extraction.}
\begin{tabular}{lcccc}
\toprule
\makecell{Substance category}       & \makecell{Precision \\(relaxed)} & \makecell{Recall \\(relaxed)} & \makecell{F1 \\(relaxed)} & \makecell{F1 \\(strict)} \\ \hline
Alcohol                           & 0.8678                       & 0.8759                    & 0.8689                & 0.8446              \\
Opioid                            & 0.8410                       & 0.8448                    & 0.8394                & 0.8106              \\
Cannabis                          & 0.8026                       & 0.8076                    & 0.8027                & 0.7805              \\
Cocaine                           & 0.7690                       & 0.7735                    & 0.7682                & 0.7401              \\
Amphetamine                       & 0.6155                       & 0.6155                    & 0.6139                & 0.5934              \\
Caffeine                          & 0.6745                       & 0.6739                    & 0.6730                & 0.6542              \\
Hallucinogen                      & 0.7794                       & 0.7792                    & 0.7768                & 0.7488              \\
Nicotine                          & 0.7971                       & 0.8018                    & 0.7975                & 0.7809              \\
Inhalant                          & 0.5931                       & 0.5889                    & 0.5887                & 0.5761              \\
Other or unknown                  & 0.7772                       & 0.7753                    & 0.7745                & 0.7571              \\
Sedative, hypnotic, or anxiolytic & 0.7324                       & 0.7354                    & 0.7317                & 0.7071             \\ \hline
Overall                           & 0.7963                       & 0.8011                    & 0.7962                & 0.7735             \\
\bottomrule
\end{tabular}
\label{tab:judge}
\end{table}

Among the triggered cases, defined as instances classified as LLM+ but ICD-negative for at least one substance category, a random sample of 100 clinical notes was reserved for SME review to evaluate the reliability of the judge LLM. The SME reviewed both the LLM-generated extractions and substance categories with no extractions. 
As shown in Table~\ref{tab:iaa}, the overall Gwet's AC1 scores and for each substance category indicated near-perfect agreement (>0.80) between the SME and the judge LLM.
Further analysis restricted to LLM extractions without corresponding ICD-10 codes showed 87.67\% agreement and a Gwet's AC1 score of 0.80.

\begin{table}[!htbp]
\centering
\caption{Inter-annotator agreement (Gwet's AC1) between the judge LLM and SME for a random sample of 100 clinical notes across substance categories.}
\label{tab:iaa}
\begin{tabular}{lcc}
\toprule
\makecell{Substance category}       & \makecell{Overall Agreement (\%)} & \makecell{Gwet's AC1} \\ \hline
Alcohol                           & 90                             & 0.87                                     \\
Opioid                            & 92                             & 0.9                                      \\
Cannabis                          & 92                             & 0.9                                      \\
Cocaine                           & 91                             & 0.88                                     \\
Amphetamine                       & 83                             & 0.79                                     \\
Caffeine                          & 95                             & 0.95                                     \\
Hallucinogen                      & 95                             & 0.95                                     \\
Nicotine                          & 95                             & 0.93                                     \\
Inhalant                          & 95                             & 0.95                                     \\
Other or unknown                  & 94                             & 0.93                                     \\
Sedative, hypnotic, or anxiolytic & 95                             & 0.95                                     \\ \hline
Overall                           & 92                             & 0.91                                     \\
\bottomrule
\end{tabular}
\end{table}

The observed level of agreement between the judge LLM and the SME suggests that the automated judge generally aligns with expert clinical interpretation when assessing whether extracted SUD severity information is accurate based on the source documentation. This alignment is notable given that the reviewed cases were drawn from a subset characterized by higher uncertainty.
Disagreements between the judge LLM and the SME likely reflect the nuances of clinical documentation, in which statements of diagnoses in different contexts can influence meaning (e.g., active vs historical diagnoses, provisional vs confirmed, etc). Such cases can reasonably support multiple interpretations and highlight known limitations of both automated and human assessment when operating on free-text clinical data. In addition, some discrepancies may arise from the judge LLM applying a more restrictive interpretation of evidentiary support than that used in expert clinical review.

For the 100 reviewed samples, we evaluated the primary LLM separately using SME annotations and judge LLM assessments as reference standards to assess potential over- or underestimation of performance by the judge LLM.
When using the judge LLM as the reference, relaxed-matching evaluation yielded a precision of 0.8179, recall of 0.8244, and F1 score of 0.8192, with an F1 score of 0.8 under strict matching. When using SME annotations as the reference standard, relaxed-matching performance resulted in a precision of 0.7828, recall of 0.7837, and F1 score of 0.7825, with a strict-matching F1 score of 0.7750.
Across both matching criteria, performance measured against SME annotations was modestly lower than performance measured against the judge LLM, likely reflecting the more conservative and comprehensive clinical interpretation applied by the human expert compared with the automated evaluator.

Error analysis indicated that most disagreements occurred in cases involving stimulant use disorders (e.g., cocaine or amphetamine), in which the judge LLM did not consistently incorporate clarifying information elsewhere in the clinical note specifying the stimulant subtype.
In some instances, the judge model appeared to apply overly conservative criteria, treating mentions of ``stimulant'' or ``other stimulant'' as insufficient supporting evidence for amphetamine use disorder.
Additional disagreements arose when the judge LLM conflated ``other or unknown substance use disorder'' with ``other stimulant use disorder''.
The model also showed limitations in interpreting documentation related to polysubstance use disorder, which clinicians may record either as a single diagnosis \cite{dsm4} or as multiple distinct substance use disorders.
Collectively, these findings highlight specific areas for targeted review and refinement of the judge model's assessments.

Within the validation framework, targeted SME review functions as a calibration mechanism rather than as a comprehensive annotation effort. The level of agreement observed supports the use of the judge LLM as a scalable confirmatory validation component, enabling efficient downstream analyses while reserving expert review for ambiguous or high-impact cases.

\subsection{Post-extraction External Validation}
\begin{figure}[!htbp]
    \centering
    \includegraphics[width=\linewidth]{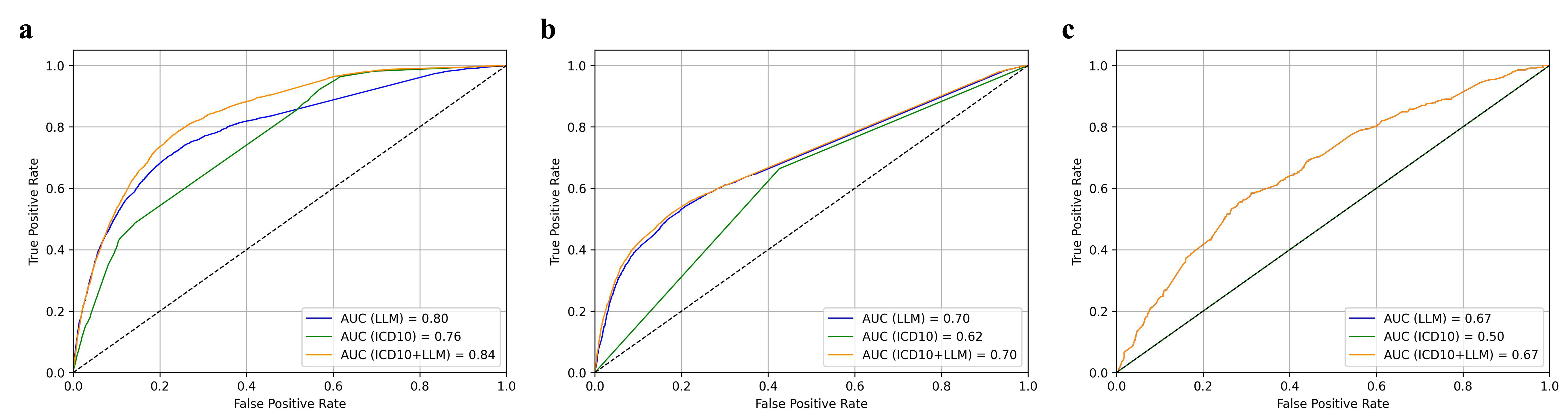}
    \caption{Predictive validity of LLM-extracted SUD diagnoses shown by ROC curves comparing outcome prediction using LLM-extracted, ICD-10–based, and combined features across cohorts (a) full study population, (b) concordant cohort, and (c) narrative-only cohort.}
    \label{fig:pred_res}
\end{figure}

External validation of LLM-derived extractions was evaluated through outcome prediction across three predefined cohorts (Fig.~\ref{fig:pred_res}). In the full study population (cohort 1), the prediction model incorporating LLM-extracted SUD diagnoses outperformed the model based on ICD-10 diagnoses alone (AUC 0.80 vs 0.76). Models combining LLM-extracted and ICD-10–based diagnoses demonstrated incremental validity, achieving the highest performance (AUC 0.84), indicating that narrative-derived extractions provide complementary predictive signal beyond structured codes. This may be due to the larger range of SUD diagnoses found in provider narratives and the ability for the provider to name the specific substance within a category as compared to the relatively coarser ICD-10 codes.

Analyses in the concordant (cohort 2) and narrative-only (cohort 3) cohorts were restricted to alcohol use disorder, resulting in reduced diagnostic breadth and sample size and, consequently, lower overall predictive performance relative to the full cohort. However, these analyses were intended for comparative evaluation rather than optimization of absolute performance. In the concordant cohort, models using LLM-extracted diagnoses outperformed those based on ICD-10 diagnoses (AUC = 0.70).
In the narrative-only cohort, LLM-extracted diagnoses achieved comparable performance (AUC = 0.67), whereas ICD-10–based diagnoses yielded an AUC of 0.50, reflecting the lack of discriminative signal due to the cohort's defining absence of structured alcohol use disorder codes.
Across both cohorts, models leveraging LLM-extracted alcohol use disorder diagnoses consistently outperformed models relying solely on ICD-10–based diagnoses. Improved performance for the alcohol use disorder cohort containing both LLM-extracted and ICD-10-based diagnoses may be due to the larger range of SUD diagnostic specifiers that can describe the severity and course of the disorder (see SUD Chapter DSM-5 APA, 2013) as compared to the relatively coarser ICD-10 codes.
These results provide external validation that LLM-derived extractions capture clinically meaningful signal aligned with future outcomes, thereby increasing confidence in the trustworthiness and reliability of LLM outputs in real-world clinical documentation settings where structured coding may be incomplete or selectively applied.

\subsection{Limitations}
This study has several limitations.
First, although the proposed validation framework is model-agnostic, our empirical evaluation was conducted using specific open-weight LLMs for extraction and confirmatory validation. While the framework is designed to generalize across instruction-tuned models, performance characteristics and error profiles may differ across model architectures, training data, and deployment settings. Future work is needed to evaluate the framework with alternative LLMs.
Second, predictive validity was assessed using future engagement with SUD specialty care as the outcome. Although this outcome reflects clinically meaningful recognition and treatment engagement, it is influenced by factors beyond clinical need, including access to care, referral patterns, and institutional practice variation. As such, predictive associations should not be interpreted causally, and additional outcome measures may be needed to fully characterize clinical impact.
Third, structured diagnosis codes were treated as imperfect indicators of SUD diagnoses and used both for cohort stratification and as a baseline comparator. Misclassification or under-coding in administrative data may therefore affect cohort definitions and performance estimates. While this limitation motivated the use of narrative extraction, residual bias arising from incomplete or inconsistent coding cannot be fully excluded.
Fourth, SME review was intentionally limited to a targeted subset of high-uncertainty cases to support scalability. Although agreement analyses provide evidence that the judge LLM aligns with expert judgment in these settings, we did not conduct exhaustive human annotation across all extractions. Consequently, unobserved error modes may persist outside the reviewed subset.
Fifth, later stages of validation depend on structured data that is closely relevant to the information of interest. When such data is unavailable, validation must be performed using indirect or inferred evidence.
Future work may assess internal consistency of LLM outputs by measuring agreement across multiple notes or encounters, with lower concordance indicating higher risk of spurious extraction. This approach could support validation in datasets without structured diagnostic labels.
Finally, although this study leveraged data from a single health system spanning 170 medical centers with documented internal variation, the proposed validation framework is not institution-specific and can be applied across settings. However, task- and site-specific adaptation of prompts for both the primary and judge LLMs, as well as adjustments to rule-based filtering components, may be required to accommodate local documentation practices. Evaluation across additional clinical settings will therefore be necessary to assess the robustness of these task-specific configurations.

\section{Conclusion}
\label{sec:conclusion}
LLMs offer substantial promise for extracting clinically meaningful information from unstructured health records, but their deployment in clinical settings is constrained by concerns about reliability and hallucination. In this work, we presented a multi-stage validation framework designed to improve the trustworthiness of LLM-based clinical information extraction by combining lightweight automated checks, targeted model-based adjudication, selective human review, and external outcome-based validation.
Our results demonstrate that each stage addresses distinct failure modes, from structurally implausible outputs and semantically unsupported content to clinically nuanced extractions that require deeper reasoning. By applying more computationally intensive validation steps only to a small, high-impact subset of uncertain cases, the framework achieves a favorable balance between rigor and scalability, enabling deployment at million-note scale without solely relying on exhaustive human annotation. Importantly, agreement between the judge LLM and subject matter experts supports the use of model-based confirmation as an efficient surrogate for expert review in targeted settings.
Beyond internal validation, predictive validity analyses show that LLM-extracted signals, particularly those derived solely from narrative documentation, exhibit associations with future clinical outcomes better than established structured indicators. This external validation provides evidence that the extracted information reflects clinically meaningful signal rather than spurious artifacts of documentation or model behavior, strengthening confidence in the use of LLM-based extraction for clinical research and decision support.
These findings suggest that trustworthy deployment of LLMs in clinical settings does not require perfect models, but rather principled validation strategies that acknowledge uncertainty, prioritize safety, and link extracted information to real-world clinical relevance. The framework presented here provides a practical path toward reliable, scalable use of LLMs for clinical information extraction.
The proposed framework is model-agnostic and can be applied to a wide range of instruction-tuned LLMs and clinical extraction tasks. While this study focused on substance use disorder diagnoses, the approach generalizes to other clinical domains characterized by incomplete or coarse structured coding and reliance on narrative documentation.
Future work will assess the applicability of the framework to other clinical information extraction tasks and evaluate its performance under prospective or temporally separated validation settings.

\printbibliography


\section*{Acknowledgment}
This initiative is sponsored by the U.S. Department of Veterans Affairs (VA) and utilizes VA-funded resources from the Knowledge Discovery Infrastructure (KDI) at Oak Ridge National Laboratory under the Department of Energy (DOE) Office of Science. The manuscript has been authored by UT-Battelle LLC under contract DE-AC05-00OR22725 with the DOE. The US government retains a nonexclusive, paid-up, irrevocable, worldwide license to publish or reproduce this manuscript or allow others to do so for US government purposes. DOE will provide public access to these results in accordance with the DOE Public Access Plan (http://energy.gov/downloads/doe-public-access-plan). The initiative aims to improve care for Veterans within the Veterans Health Administration (VHA) and involves models specifically tailored to the VHA system. The results are not generalizable outside of VHA, and the studies are formally considered non-research by the VHA. Oak Ridge National Laboratory received Institutional Review Board (IRB) approval for the secondary use of patient data, which complies with internal policies and standards. The authors acknowledge the broader partnership and express gratitude to the Veterans receiving care at the VA.



\end{document}